\documentclass[sigconf]{acmart}
\AtBeginDocument{%
  }

\copyrightyear{2026}
\acmYear{2026}
\setcopyright{cc}
\setcctype{by}
\acmConference[MM '26]{Proceedings of the 34th ACM International Conference on Multimedia}{November 10--14, 2026}{Rio de Janeiro, Brazil}
\acmBooktitle{Proceedings of the 34th ACM International Conference on Multimedia (MM '26), November 10--14, 2026, Rio de Janeiro, Brazil}
\acmDOI{10.1145/3767308.3835433}
\acmISBN{979-8-4007-2213-4/2026/11}

\usepackage{booktabs}  
\usepackage{colortbl}  
\usepackage{xcolor}    

\usepackage{amssymb}   
\usepackage{graphicx}  
\usepackage{multirow}  
\usepackage{pifont}
\newcommand{\cmark}{\ding{51}}
\newcommand{\xmark}{\ding{55}}

\begin{document}

\title{MuSS: A Large-Scale Dataset and Cinematic Narrative Benchmark for Multi-Shot Subject-to-Video Generation}
\titlenote{Project page: \url{https://github.com/zhang-haojie/MuSS}}


\author{Haojie Zhang}
\authornote{Both authors contributed equally to this research.}
\affiliation{%
  \institution{South China University of Technology}
  \city{Guangzhou}
  \country{China}}
\email{eehaojiezhang@gmail.com}

\author{Di Wu}
\authornotemark[1]
\affiliation{%
  \institution{South China University of Technology}
  \city{Guangzhou}
  \country{China}}
\email{qiuyi900800@gmail.com}

\author{Bingyan Liu}
\affiliation{%
  \institution{South China University of Technology}
  \city{Guangzhou}
  \country{China}}
\email{liubingyan.lby@gmail.com}

\author{Linjie Zhong}
\affiliation{%
  \institution{South China University of Technology}
  \city{Guangzhou}
  \country{China}}
\email{eelinjie@mail.scut.edu.cn}

\author{Yuancheng Wei}
\affiliation{%
  \institution{Peking University}
  \city{Beijing}
  \country{China}}
\email{2601112128@stu.pku.edu.cn}

\author{Xingsong Ye}
\affiliation{%
  \institution{Fudan University}
  \city{Shanghai}
  \country{China}}
\email{xsye20@fudan.edu.cn}

\author{Nanqing Liu}
\authornote{Corresponding author.}
\affiliation{%
  \institution{Yunnan Normal University}
  \city{Kunming}
  \country{China}}
\email{lansing163@163.com}

\author{Yaling Liang}
\affiliation{%
  \institution{South China University of Technology}
  \city{Guangzhou}
  \country{China}}
\email{ylliang@scut.edu.cn}

\renewcommand{\shortauthors}{Zhang et al.}
\hypersetup{
  pdftitle={MuSS: A Large-Scale Dataset and Cinematic Narrative Benchmark for Multi-Shot Subject-to-Video Generation},
  pdfauthor={Haojie Zhang, Di Wu, Bingyan Liu, Linjie Zhong, Yuancheng Wei, Xingsong Ye, Nanqing Liu, and Yaling Liang}
}

\begin{abstract}
While video foundation models excel at single-shot generation, real-world cinematic storytelling inherently relies on complex multi-shot sequencing. Further progress is constrained by the absence of datasets that address three core challenges: authentic narrative logic, spatiotemporal text-video alignment conflicts, and the "copy-paste" dilemma prevalent in Subject-to-Video (S2V) generation. To bridge this gap, we introduce MuSS, a large-scale, dual-track dataset tailored for multi-shot video and S2V generation. Sourced from over 3,000 movies, MuSS explicitly supports both complex montage transitions and subject-centric narratives. To construct this dataset, we pioneer a progressive captioning pipeline that eliminates contextual conflicts by ensuring local shot-level accuracy before enforcing global narrative coherence. Crucially, we implement a cross-shot matching mechanism to fundamentally eradicate the S2V copy-paste shortcut. Alongside the dataset, we propose the Cinematic Narrative Benchmark, featuring a visual-logic-driven paradigm and a novel Anti-Copy-Paste Variance (ACP-Var) metric to rigorously assess continuous storytelling and 3D structural consistency. Extensive experiments demonstrate that while current baselines struggle with continuous narrative logic or degenerate into trivial 2D sticker generators, our MuSS-augmented model achieves state-of-the-art narrative effectiveness and cross-shot identity preservation.
\end{abstract}


\begin{teaserfigure}
  \centering
  \includegraphics[width=0.95\textwidth]{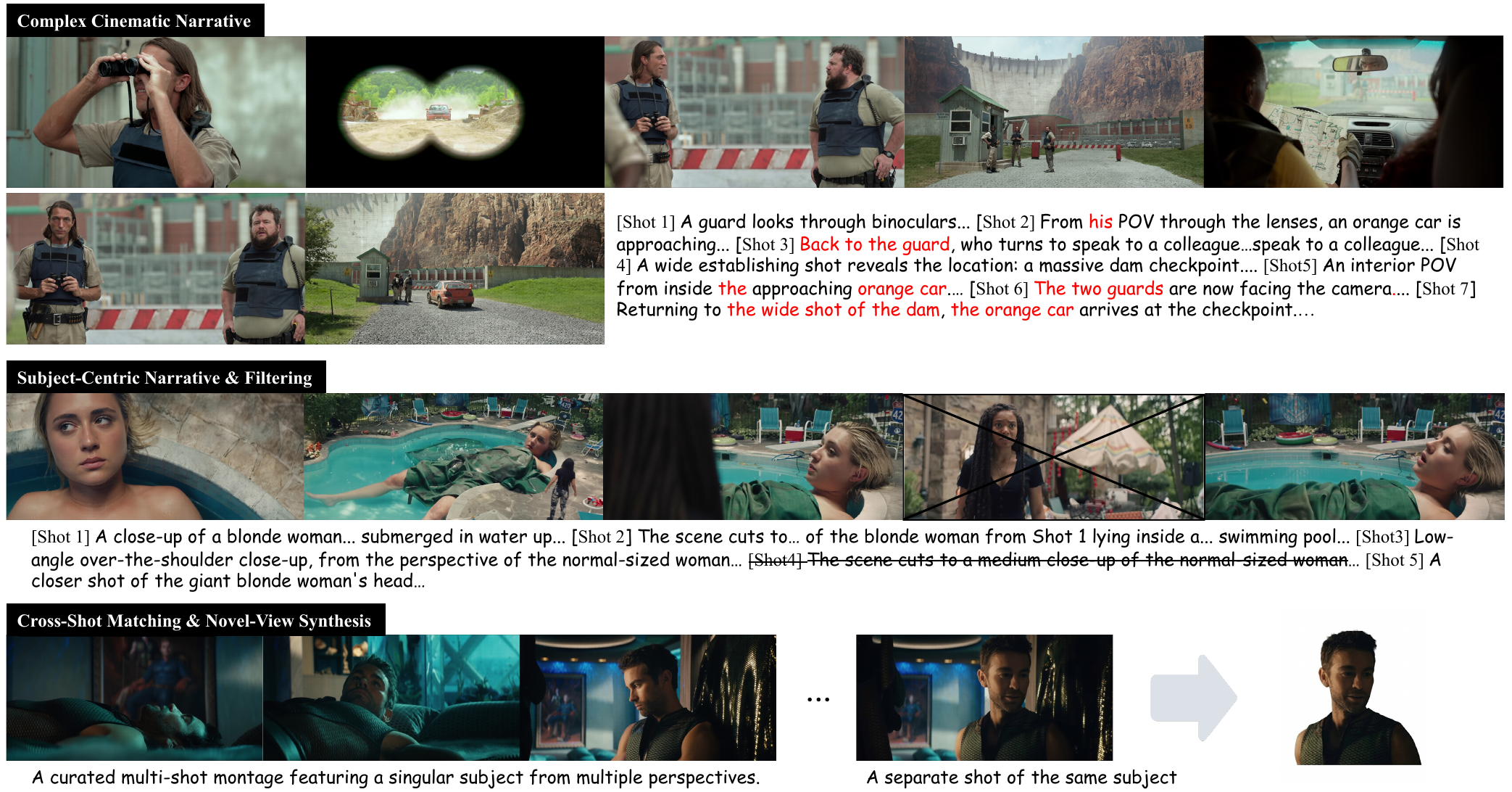}
  \caption{Overview of the MuSS dataset construction. (Top) Complex Cinematic Narrative: Progressive captioning resolves cross-shot coreferences (in red) for precise narrative alignment. (Middle) Subject-Centric Narrative: Intervening shots are filtered to maintain continuous focus on the core identity. (Bottom) Cross-Shot Matching: To break the "copy-paste" shortcut in S2V generation, the reference subject is extracted from a separate shot, forcing models to learn novel-view synthesis.}
  \Description{A three-row pipeline diagram. The top row shows progressive captioning of cinematic shots with resolved cross-shot entity references. The middle row shows construction of subject-centric narratives after filtering intervening shots. The bottom row shows a subject reference extracted from a disjoint shot and paired with a target multi-shot video.}
  \label{fig:teaser}
\end{teaserfigure}


\maketitle

\section{Introduction}

Diffusion Models have rapidly advanced Text-to-Video (T2V) and Subject-to-Video (S2V) generation \cite{guo2023animatediff, brooks2024video, kong2024hunyuanvideo, wan2025wan, zhou2024storydiffusion, zhao2024moviedreamer}. However, existing open-source datasets (e.g., OpenS2V-5M \cite{yuan2025opens2v}) and generation frameworks are predominantly confined to isolated, single-shot videos, typically focusing on simple actions of one subject. Cinematic production, advertising, and short-form content instead rely on complex multi-shot sequencing \cite{xiao2025captain, meng2025holocine, wu2025cinetrans, kara2025shotadapter, wang2025multishotmaster, cai2025mixture, he2025cut2next}. Transitions between diverse subjects and scenes drive the narrative. Consequently, the scarcity of multi-shot datasets encapsulating authentic cinematic language is a primary bottleneck to industrial-grade video generation.

Constructing a high-quality multi-shot dataset poses three challenges. \textbf{(1) Scarcity of Real Narrative Logic:} Movies feature intricate camera blocking and montage (e.g., transitioning from an establishing shot to Subject A's close-up, then to Subject B). Concatenating independent single-shot videos fails to simulate this narrative structure. \textbf{(2) Spatiotemporal Text Alignment and Conflict:} In multi-subject or multi-scene transitions, global captioning struggles with fine-grained control over individual shots, whereas independent captions can become contradictory when merged into a sequence. \textbf{(3) The ``Copy-Paste'' Dilemma in S2V Generation:} Cinematic storytelling requires consistent subjects across varied viewpoints. Customized S2V generation builds on image personalization techniques such as DreamBooth \cite{ruiz2023dreambooth} and IP-Adapter \cite{ye2023ip}. However, if the reference subject is extracted from the target frame, existing models \cite{mao2024story, yuan2025identity, wang2025echoshot} often replicate its pose and lighting. This degrades generalization to novel views across shots.

To overcome these challenges, we introduce \textbf{MuSS}, a large-scale, open-source dataset for multi-shot video and S2V generation (Figure \ref{fig:teaser}). Sourced from over 3,000 movies, it comprises approximately 700K filtered shots, from which more than 30K captioned multi-shot sequences are constructed after multi-dimensional filtering (e.g., aesthetics \cite{zhai2023sigmoid}, motion \cite{teed2020raft}, and semantic consistency \cite{radford2021learning, caron2021emerging}).
Unlike datasets predominantly confined to isolated subjects, MuSS covers two real-world narrative settings: \textit{(i) Complex Cinematic Narrative}, involving montage transitions between subjects and scenes within one storyline; and \textit{(ii) Subject-Centric Narrative}, generating shots for the same identity across scenes and timelines.
This dual-track composition forms a holistic storytelling solution: the first track teaches the structural logic of narrative editing, while the second promotes 3D identity preservation under perspective shifts. Together, they address the limitations compared in Table \ref{tab:dataset_comparison}.

Existing benchmarks such as VBench \cite{huang2024vbench}, EvalCrafter \cite{liu2024evalcrafter}, MSVBench \cite{shi2026msvbench}, and ViStoryBench \cite{zhuang2025vistorybench} focus on global video quality and basic textual alignment, but fall short in evaluating the spatiotemporal logic required for storytelling. We propose the \textbf{Cinematic Narrative Benchmark}, a dual-track suite for realistic storytelling conditions.
First, for \textit{Narrative Effectiveness Validation} (targeting complex cinematic narratives), we assess the model's storytelling ability across multi-subject and multi-view transitions. We employ \textbf{Structural Text Alignment} to ensure each physical shot precisely matches its local prompt without semantic bleeding, alongside \textbf{Multi-Shot Temporal Coherence} to measure the naturalness of transitions.
Second, for \textit{Subject Consistency Validation} (S2V settings), we evaluate cross-shot identity preservation. Beyond \textbf{Face/ID Preservation}, we introduce \textbf{Anti-Copy-Paste Variance (ACP-Var)}. By quantifying structural and pose diversity between the reference and generated videos, it tests novel-view generation rather than shortcut memorization.

\begin{table*}[t]
\centering
\caption{Comparison with existing video generation datasets. Unlike most datasets, MuSS supports multi-shot subject-to-video generation with high-resolution cinematic clips.}
\label{tab:dataset_comparison}
\resizebox{\textwidth}{!}{
\begin{tabular}{l c c c c c c c}
\toprule
\textbf{Dataset} & \textbf{Multi-shot} & \textbf{Subject-to-Video} & \textbf{Resolution} & \textbf{Video Clips} & \textbf{Video Duration (h)} & \textbf{Open-Source} & \textbf{Type} \\
\midrule
MSRVTT \cite{xu2016msr}            & \xmark & \xmark & 240P  & 10K   & 40   & \cmark & Web Videos \\
WebVid-10M \cite{bain2021frozen}         & \xmark & \xmark & 360P  & 10M   & 52K  & \cmark & Web Videos \\
OpenVid-1M \cite{nan2024openvid}        & \xmark & \xmark & Diverse& 1M   & 2K   & \cmark & Web Videos \\
InternVid \cite{wang2023internvid}       & \xmark & \xmark & 720P  & 234M  & 760K & \cmark & Web Videos \\
LVD-2M \cite{xiong2024lvd}              & \xmark & \xmark & Diverse& 2M   & -    & \cmark & Web Videos \\
OpenHumanVid \cite{li2025openhumanvid} & \xmark & \xmark & 720P  & 52.3M & 70K  & \cmark & Human Videos \\
VBench \cite{huang2024vbench}             & \xmark & \xmark & Various& 100K+& -    & \cmark & Synthetic / Web \\
OpenS2V-5M \cite{yuan2025opens2v}  & \xmark & \cmark & 720P  & 5.4M  & 10K  & \cmark & Web Videos \\
\midrule
LLaVA-Video \cite{zhang2024llava}    & \cmark & \xmark & Diverse& 178K & 2K   & \cmark & Web Videos \\
Shot2Story \cite{han2023shot2story20k}     & \cmark & \xmark & 720P  & 134K  & -    & \cmark & Web Videos \\
Cine250K \cite{wu2025cinetrans}  & \cmark & \xmark & 720P  & 250K  & 740  & \cmark & Movies \\
PortraitGala \cite{wang2025echoshot} & \cmark & \cmark & Various& 400K & 1K   & \xmark & Human Portraits \\
\midrule
\rowcolor{gray!20} 
\textbf{MuSS (Ours)}         & \textbf{\cmark} & \textbf{\cmark} & \textbf{720P} & \textbf{700K} & \textbf{1K} & \textbf{\cmark} & \textbf{Movies} \\
\bottomrule
\end{tabular}
}
\end{table*}

In summary, our main contributions are as follows:
\begin{itemize}
\item We construct MuSS, a large-scale, high-quality multi-shot video library derived from cinematic material.
\item We introduce progressive VLM annotation and precise cross-shot subject matching. Using subjects from alternate shots as references encourages natural novel views and prevents the ``copy-paste'' shortcut.
\item We propose the Cinematic Narrative Benchmark, replacing coarse global text evaluation with a Visual-Logic driven paradigm. Multi-Dimensional Visual Logic and Anti-Copy-Paste Variance (ACP-Var) expose structural hallucination and 2D sticker generation.
\item Experiments show that while current baselines struggle with cinematic multi-shot scenarios, our MuSS-augmented baseline achieves state-of-the-art storytelling effectiveness, structural grounding, and identity consistency.
\end{itemize}

\section{Related Work}

\subsection{Multi-Shot and Long Video Generation}
Coherent long-video generation has evolved from simple temporal extrapolation to complex narrative modeling. StoryDiffusion \cite{zhou2024storydiffusion} introduced consistent self-attention for long-range generation, while MovieDreamer \cite{zhao2024moviedreamer} proposed a hierarchy for coherent visual sequences. For long-duration talking videos, LetsTalk \cite{zhang2024letstalk} employs a noise-regularized memory bank to maintain contextual continuity and mitigate error accumulation during extended generation.
Recent work focuses on authentic cinematic storytelling and multi-shot coherence. These frameworks include CineTrans \cite{wu2025cinetrans}, which uses masked diffusion models for cinematic transitions, ShotAdapter \cite{kara2025shotadapter}, and MultiShotMaster \cite{wang2025multishotmaster}. Captain Cinema \cite{xiao2025captain} and HoloCine \cite{meng2025holocine} attempt to generate complete short-film narratives. Managing long-context dependencies has inspired in-context solutions such as Mixture of Contexts \cite{cai2025mixture}, Long Context Tuning \cite{guo2025long}, MoGA \cite{jia2025moga}, and Cut2Next \cite{he2025cut2next}.
New benchmarks and datasets include MSVBench \cite{shi2026msvbench}, ViStoryBench \cite{zhuang2025vistorybench}, and domain-specific AnimeShooter \cite{qiu2025animeshooter} and FairyGen \cite{zheng2025fairygen}. Despite these commendable efforts, existing datasets often lack the real-world cinematic logic and complex scene transitions required for industrial-grade multi-shot generation.

\subsection{Subject-to-Video Generation}
Maintaining strict identity (ID) across varying views and scenes is the core challenge of customized generation. Image-level techniques such as WithAnyone \cite{xu2025withanyone}, MultiRef \cite{chen2025multiref}, and OpenSubject \cite{liu2025opensubject} provide spatial priors that have been extended to video.
Recent video models achieve impressive zero-shot identity preservation. Magic Mirror \cite{zhang2025magicmirror} uses video diffusion transformers, while Phantom \cite{liu2025phantom} uses cross-modal alignment for subject consistency. Kaleido \cite{zhang2025kaleido} extends reference video generation to multiple subjects. EchoShot \cite{wang2025echoshot} specifically targets multi-shot portrait videos, and related work highlights the initial frame's critical role in content customization \cite{yuan2025opens2v}.
Large-scale resources such as OpenS2V-Nexus \cite{yuan2025opens2v} standardize evaluation. However, existing S2V datasets focus mainly on isolated, single-shot actions and can encourage the ``copy-paste'' shortcut. Consequently, they fail to rigorously benchmark 3D identity preservation across dynamic, multi-shot cinematic transitions.

\section{MuSS Dataset Construction}
\label{sec:dataset}

\begin{figure*}
    \centering
    \includegraphics[width=0.95\linewidth]{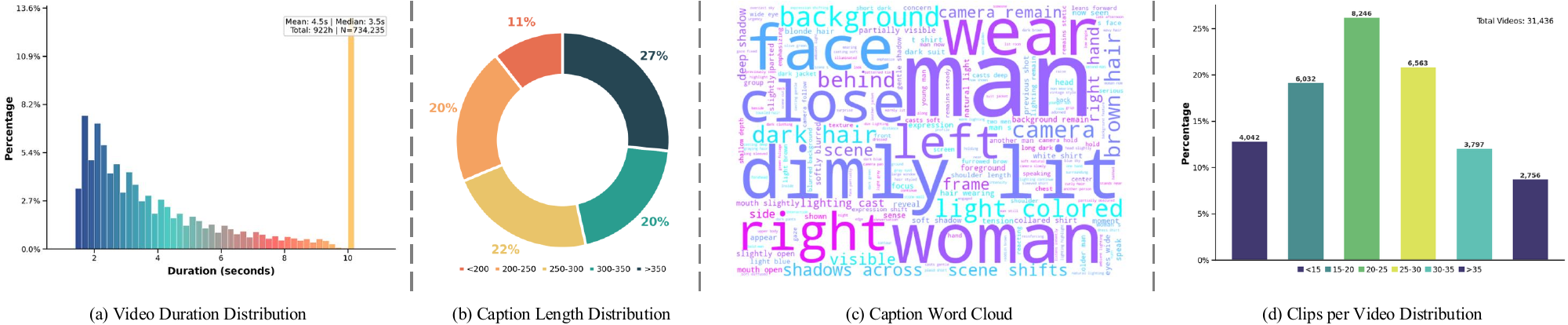}
    \caption{Overview of the MuSS dataset statistics. (a) Video clip duration distribution. (b) Caption length distribution. (c) Caption word cloud. (d) Number of clips per source video.}
    \Description{Four dataset statistics panels showing distributions of video-clip duration and caption length, a word cloud of common caption terms, and the number of retained clips per source movie.}
    \label{fig:statistics} 
\end{figure*}

\begin{figure*}[t]
\centering
\includegraphics[width=0.95\linewidth]{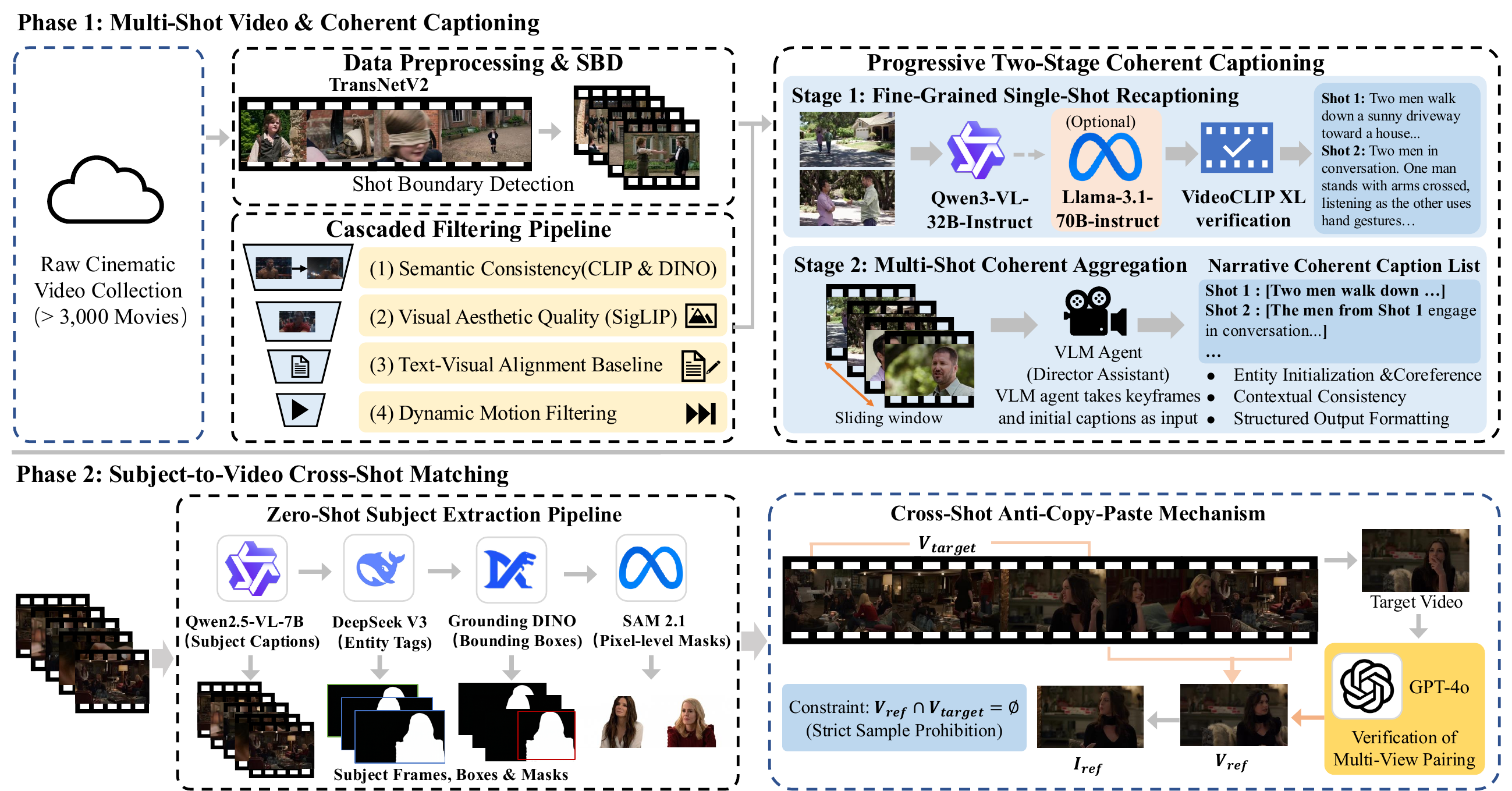}
\caption{Illustration of the MuSS dataset curation methodology. (a) Multi-Shot Video and Coherent Captioning: Transforms unconstrained cinematic footage into structured multi-shot clips through cascaded filtering and a two-stage VLM recaptioning pipeline. (b) Cross-Shot Matching for S2V: Employs zero-shot subject-centric extraction and explicitly samples reference images from disjoint shot contexts to construct a robust customized generation benchmark.}
\Description{A two-part flowchart. The first part decomposes movies into shots, applies cascaded quality filters, groups neighboring shots, and generates coherent local and global captions. The second part detects and tracks subjects, extracts subject crops, and pairs references from disjoint shots with target videos.}
\label{fig:dataset_pipeline}
\end{figure*}

We construct \textbf{MuSS}, a large-scale dataset for multi-shot generation and evaluation. Over 3,000 movies yield approximately 700K filtered single shots, more than 30,000 professionally captioned multi-shot clips, and over 1,000 hours of high-quality video (Figure \ref{fig:statistics}). Construction has two phases (Figure \ref{fig:dataset_pipeline}): (1) building a high-quality multi-shot foundation with coherent text alignment, and (2) curating Subject-to-Video (S2V) pairs through cross-shot matching to prevent the ``copy-paste'' shortcut.

\subsection{Multi-Shot Video and Coherent Captioning}
The first phase transforms raw movies into high-quality multi-shot sequences paired with narrative-coherent captions.

\noindent\textbf{Data Preprocessing and Shot Boundary Detection.} 
Raw videos are preprocessed by removing watermarks and cropping black borders (letterboxing/pillarboxing). We then employ TransNetV2 \cite{soucek2024transnet} for Shot Boundary Detection (SBD). Its temporal representation handles hard cuts and gradual transitions such as fades and dissolves, ensuring that each clip contains a single, continuous camera shot.

\noindent\textbf{Multi-Dimensional Cascaded Filtering Pipeline.} 
Raw shots often contain motion blur, static scenes, or meaningless transitions. We apply a cascaded filter to select training candidates. Importantly, filtering occurs before sliding-window sequence construction and never removes shots from an assembled narrative sequence:
\textit{Semantic Consistency:} We utilize CLIP \cite{radford2021learning} and DINO \cite{caron2021emerging} to compute the semantic similarity between the keyframe and the first frame of each shot. Shots demonstrating insufficient semantic consistency are discarded to ensure intra-shot stability and rule out abrupt visual shifts.
\textit{Visual Aesthetic Quality:} We employ the SigLIP \cite{zhai2023sigmoid} model to evaluate the aesthetic score of uniformly sampled frames, retaining only those that meet a high cinematic visual standard.
\textit{Text-Visual Alignment Baseline:} A preliminary text score filter is applied to remove clips that completely lack semantic describability or meaningful visual concepts.
\textit{Dynamic Motion Filtering:} Cinematic videos must exhibit appropriate dynamics. We compute a motion score for each shot and restrict it within a reasonable range. This effectively filters out frozen content while retaining low-motion dialogue and establishing shots, and also excludes excessively chaotic camera movements that could disrupt the latent space of video diffusion models.

\noindent\textbf{Progressive Two-Stage Coherent Captioning.} 
The key challenge is aligning descriptions with physical shots without contextual conflict. We adopt a ``single-shot first, multi-shot second'' progressive Vision-Language Model (VLM) annotation pipeline.

\textit{Stage 1: Fine-Grained Single-Shot Recaptioning.} 
Instead of coarse metadata, we deploy Qwen3-VL-32B-Instruct \cite{bai2025qwen3} for fine-grained independent shot descriptions, optionally utilizing Llama-3.1-70B-Instruct \cite{grattafiori2024llama} to rewrite captions for prompt-friendliness.
Finally, we compute the VideoCLIPXL \cite{xu2021videoclip} score between the rewritten caption and the video clip, discarding any pairs with an alignment score below $0.20$.

\textit{Stage 2: Multi-Shot Coherent Aggregation.} 
To construct narrative multi-shot sequences, we apply a sliding window approach over the consecutive single shots. To aggregate these shots into a cohesive storyline, we design a specialized VLM agent acting as a ``film-director assistant''. The VLM takes the keyframes and initial single-shot captions of the sequence as input and globally refines them under strict narrative constraints: (1) \textbf{Entity Initialization and Coreference:} Characters or objects are explicitly introduced only upon their first appearance, and referred to using consistent pronouns in subsequent shots to avoid redundancy. (2) \textbf{Contextual Consistency:} The VLM ensures logical flow and eliminates contradictory descriptions of the same subject across different views. (3) \textbf{Structured Formatting:} The VLM outputs precisely structured text strictly aligned with the physical shot count (e.g., ``\texttt{Shot 1: [caption] \textbackslash n ...}''). This paradigm guarantees that the final multi-shot captions possess both frame-level control accuracy and profound cinematic narrative coherence.

\subsection{Cross-Shot Matching for Subject-to-Video Generation}
Constructing high-quality Subject-to-Video (S2V) pairs requires precise identity extraction and strategic reference sampling to prevent models from falling into the ``copy-paste'' shortcut. We develop a zero-shot subject extraction pipeline followed by a cross-shot matching mechanism to ensure 3D identity consistency.

\noindent\textbf{Zero-Shot Subject-Centric Extraction.} 
To decouple a subject's 3D identity from complex cinematic backgrounds, we design an automated perception pipeline. We first prompt Qwen2.5-VL-7B \cite{bai2025qwen25vl} for subject-centric captions and employ DeepSeekV3 \cite{liu2024deepseek} to extract concise entity tags. These tags guide GroundingDINO \cite{liu2024grounding} to detect objects in the initial frame, providing bounding boxes for Segment Anything Model 2.1 (SAM 2.1) \cite{ravi2024sam} to generate pixel-level masks. To ensure scientific rigor, we incorporate a temporal mask-consistency check to mitigate failures caused by occlusion or motion blur. This process isolates high-fidelity subject representations, forcing the model to prioritize core identity features over background layouts.

\noindent\textbf{Cross-Shot Anti-Copy-Paste Mechanism.} 
Standard S2V datasets typically sample reference images directly from the target video, leading models to learn trivial mappings of pose and lighting rather than true identity. To eradicate this shortcut, we introduce the Cross-Shot Matching Mechanism. Let $S = \{V_1, V_2, \dots, V_N\}$ denote a continuous cinematic storyline. For a target clip $V_{\text{target}} \in S$, we explicitly prohibit sampling the reference image $I_{\text{ref}}$ from $V_{\text{target}}$. Instead, we use cross-video tracking to identify the same subject in a disjoint context $V_{\text{ref}} \in S$ outside the shots that constitute the target multi-shot clip. To ensure absolute context isolation, we enforce a strict temporal displacement: $V_{\text{ref}}$ and $V_{\text{target}}$ must be separated by at least $K=1$ intervening shot or a minimum of 32 frames. Additionally, we utilize GPT-4o \cite{hurst2024gpt4o} to verify cross-frame pairings and maximize multi-view diversity. This spatial and temporal displacement ensures significant variance in camera angles and poses between the reference and target, compelling the S2V model to learn robust 3D structural comprehension and novel-view synthesis.

\section{Cinematic Narrative Benchmark} 
\label{sec:benchmark}

Existing video generation benchmarks focus on coarse global assessment of single-shot videos \cite{huang2024vbench, liu2024evalcrafter}. They do not adequately measure storytelling, cross-shot visual stability, or spatiotemporal control. We therefore propose the \textbf{Cinematic Narrative Benchmark}, a dual-track suite derived from MuSS.

\begin{figure*}[t]
\centering
\includegraphics[width=0.9\linewidth]{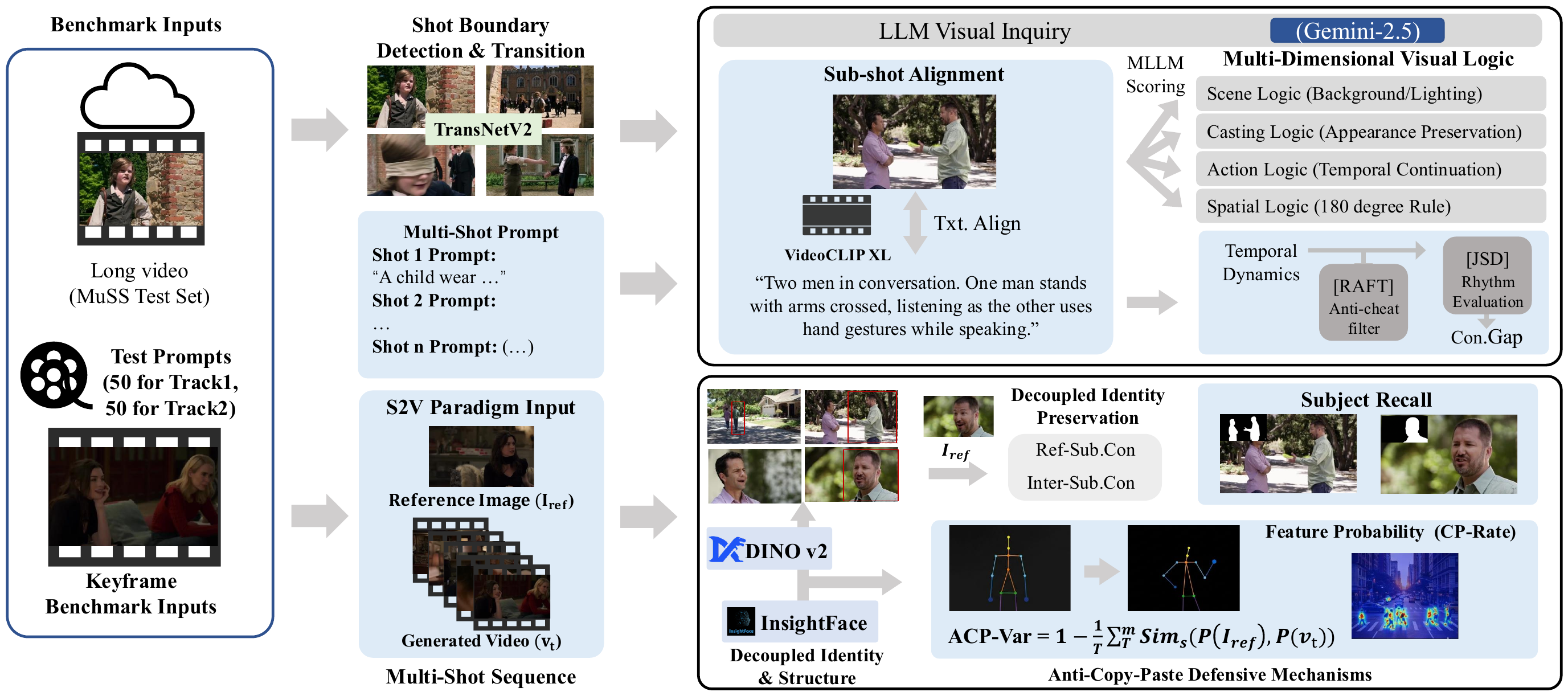}
\caption{Overview of the Cinematic Narrative Benchmark. The evaluation suite employs a novel Visual-Logic Driven paradigm, utilizing Large Multimodal Models (LMMs) and domain-expert models across two tracks. Track 1 assesses global narrative effectiveness and cross-shot visual stability, while Track 2 evaluates subject consistency and explicitly penalizes trivial "copy-paste" behaviors in Subject-to-Video (S2V) generation.}
\Description{A benchmark flowchart with two tracks. Track 1 evaluates text alignment, transition behavior, scene continuity, and multidimensional visual logic. Track 2 compares a reference subject with generated shots to measure reference fidelity, inter-shot identity consistency, pose variation, recall, and copy-paste behavior.}
\label{fig:bench_pipeline}
\end{figure*}

Because perfect global captions are impractical at scale, our benchmark adopts a \textit{Visual-Logic Driven} evaluation paradigm. As illustrated in Figure \ref{fig:bench_pipeline}, it synergizes the pure visual reasoning capabilities of Gemini-2.5-Flash \cite{geminiteam2025gemini25} with the perceptual fidelity of domain-specific expert models (e.g., DINOv2 \cite{oquab2023dinov2}, TransNet V2 \cite{soucek2024transnet}, RAFT \cite{teed2020raft}, YOLOv11 \cite{yolo11_ultralytics}, SAM \cite{ravi2024sam}). This synergistic approach allows us to achieve human-level precision in structural assessment without relying on generic global text priors.

\subsection{Track 1: Narrative Effectiveness Validation} 
Track 1 evaluates whether models follow local shot instructions while preserving the global visual world along three dimensions:

\noindent \textbf{Sub-shot Text Alignment \& Transition Precision:} Instead of a global CLIP score that masks cross-shot prompt bleeding, we compute the average VideoCLIP score between each physical shot and its local prompt (\textbf{Txt.Align}). While VideoCLIP provides a quantitative baseline, we heavily incorporate LMM visual logic to avoid unfairly penalizing valid cinematic choices (e.g., an over-the-shoulder shot temporarily omitting a subject). Furthermore, to explicitly assess multi-shot temporal controllability, we measure the transition timestamp deviation (\textbf{Trans.Dev}) using TransNet V2 for accurate boundary detection.

\noindent \textbf{Multi-Dimensional Visual Logic (MDVL) \& Scene Consistency:} We upgrade the traditional single-score LMM evaluation into a rigorous \textbf{MDVL} framework. This suite assesses generated sequences across four specific axes: \textit{Scene Logic} (stability of background and lighting after cuts), \textit{Casting Logic} (appearance preservation of the ensemble cast, deliberately designed to tolerate valid perspective shifts), \textit{Action Logic} (temporal continuation of dynamic behaviors), and \textit{Spatial Logic} (adherence to cinematic rules like the 180-degree axis). The reported \textbf{Scene.Logic} score is produced by Gemini-2.5-Flash. This LMM evaluation is complemented by \textbf{Scene.Con}, an auxiliary objective diagnostic that calculates DINOv2 similarity between SAM-cropped backgrounds across different shots; Scene.Con is not reported as an MDVL dimension and is not used interchangeably with Scene.Logic.

\noindent \textbf{Temporal Dynamics \& Consistency Gap:} To prevent models from cheating spatial consistency metrics by generating static ``slideshows,'' we utilize RAFT to quantify motion magnitude, effectively filtering out generations lacking necessary temporal dynamics. For the remaining valid videos, we compute the Jensen-Shannon Distance (JSD) between their coherence distribution and a reference set of professional film edits, yielding the \textbf{Con.Gap} metric to evaluate authentic narrative rhythm.

\subsection{Track 2: Subject Consistency Validation} 
The second track targets the Subject-to-Video (S2V) generation paradigm, assessing whether models have acquired true 3D structural consistency rather than resorting to trivial 2D pixel copying.

\noindent \textbf{Decoupled Identity Preservation:} To explicitly expose shortcut learning, we decouple the evaluation of identity preservation. \textbf{Ref-Sub.Con} evaluates the generation's fidelity to the external 2D reference image, while \textbf{Inter-Sub.Con} strictly measures internal identity preservation across generated multi-shot sequences using DINOv2 and InsightFace \cite{deng2019arcface}. Additionally, we replace the rigid cross-shot mIoU, which incorrectly penalizes valid perspective shifts like wide-to-closeup, with a more robust \textbf{Subject Recall} metric. By utilizing YOLOv11 as an out-of-the-box object detector, we verify the reliable presence of the target subject within designated frames, measuring authentic spatiotemporal grounding without punishing legitimate cinematic camera changes.

\noindent \textbf{Anti-Shortcut Metrics \& Dynamics:} To explicitly penalize models that act as mere ``2D sticker generators,'' we introduce \textbf{ACP-Var} to quantify the structural and pose diversity between the reference image $I_{\text{ref}}$ and the generated frames $v_t$:
\begin{equation}
\text{ACP-Var} = 1 - \frac{1}{T} \sum_{t=1}^{T} \text{Sim}_{\text{pose}} \left( \mathcal{P}(I_{\text{ref}}), \mathcal{P}(v_t) \right)
\end{equation}
where $\mathcal{P}(\cdot)$ extracts 2D keypoints via DWPose \cite{yang2023effective}. The $\text{Sim}_{\text{pose}}$ evaluates the cosine similarity of Procrustes-aligned keypoints, explicitly penalizing generations that lazily collapse into the rigid 2D posture of the reference image. Complementarily, \textbf{Copy-Paste Rate (CP-Rate)} detects appearance overfitting by computing the Softmax entropy of DINOv2 \cite{oquab2023dinov2} feature similarities against the reference. A near-zero entropy exposes trivial 2D duplication. Finally, \textbf{Action Strength (Act.Str)} evaluates temporal dynamics using the average RAFT \cite{teed2020raft} optical flow magnitude, effectively penalizing static generations.

\subsection{Evaluation Protocol} 
We manually verify 100 prompts from the MuSS test set, with 50 per track and stratified coverage of shot count, scene type, dialogue/action rhythm, framing, and subject count. We evaluate storyboard pipelines \cite{zhou2024storydiffusion, wan2025wan}, native multi-shot models \cite{wu2025cinetrans, meng2025holocine, wang2025echoshot}, and customized S2V frameworks \cite{liu2025phantom, jiang2025vace}, together with a physical copy-paste baseline for validating the anti-shortcut metrics.

\begin{table*}[t]
\centering
\setlength{\tabcolsep}{4.5pt}
\caption{Quantitative Results on Track 1 (Narrative Effectiveness). Our MuSS-augmented baseline sweeps all multi-dimensional visual logic metrics, maintaining rigorous consistency across continuous multi-shot storytelling while achieving highly competitive precision in spatiotemporal transitions.}
\label{tab:track1}
\begin{tabular}{l c c c c c c c}
\toprule
\multirow{2}{*}{\textbf{Method}} & \multicolumn{2}{c}{\textbf{Alignment \& Flow}} & \multicolumn{4}{c}{\textbf{Multi-Dimensional Visual Logic}} & \textbf{Dynamics} \\
\cmidrule(lr){2-3} \cmidrule(lr){4-7} \cmidrule(lr){8-8}
& \textbf{Txt.Align} $\uparrow$ & \textbf{Trans.Dev} $\downarrow$ & \textbf{Scene.Logic} $\uparrow$ & \textbf{Casting.Logic} $\uparrow$ & \textbf{Act.Logic} $\uparrow$ & \textbf{Spat.Logic} $\uparrow$ & \textbf{Con.Gap} $\downarrow$ \\
\midrule
StoryDiff \cite{zhou2024storydiffusion} + I2V \cite{wan2025wan} & 0.2260 & -    & 3.30 & 3.65 & 2.65 & 2.78 & \textbf{0.3020} \\
EchoShot \cite{wang2025echoshot}        & 0.1497 & 2.65 & 3.29 & 3.62 & 2.71 & 2.83 & 0.4080 \\
CineTrans \cite{wu2025cinetrans}       & 0.1382 & 5.43 & 3.38 & 2.88 & 2.67 & 2.92 & 0.3841 \\
HoloCine \cite{meng2025holocine}        & 0.2292 & \textbf{2.50} & 3.67 & 3.42 & 2.96 & 2.88 & 0.3631 \\
\midrule
\rowcolor{gray!15} 
\textbf{MuSS (Ours)} & \textbf{0.2359} & 2.55 & \textbf{3.84} & \textbf{3.96} & \textbf{3.12} & \textbf{3.05} & 0.3560 \\
\bottomrule
\end{tabular}
\end{table*}

\section{Experiments}
\label{sec:experiments}

We evaluate MuSS and the Cinematic Narrative Benchmark quantitatively and qualitatively.

\begin{table*}[t]
\centering
\caption{Quantitative Results on Track 2 (Subject Consistency). By decoupling reference fidelity and inter-shot consistency, we expose the structural fragility of existing S2V models. Our method uniquely breaks the copy-paste shortcut while achieving state-of-the-art grounding and internal identity preservation among customizable baselines.}
\label{tab:track2}
\begin{tabular}{l c c c c c c}
\toprule
\multirow{2}{*}{\textbf{Method}} & \multicolumn{2}{c}{\textbf{Identity Preservation}} & \multicolumn{2}{c}{\textbf{Subject Grounding}} & \multicolumn{2}{c}{\textbf{Anti-Copy-Paste}} \\
\cmidrule(lr){2-3} \cmidrule(lr){4-5} \cmidrule(lr){6-7}
& \textbf{Ref-Sub.Con (\%)} $\uparrow$ & \textbf{Inter-Sub.Con (\%)} $\uparrow$ & \textbf{Subj.Recall} $\uparrow$ & \textbf{Act.Str} $\uparrow$ & \textbf{ACP-Var} $\uparrow$ & \textbf{CP-Rate (\%)} $\downarrow$ \\
\midrule
StoryDiff \cite{zhou2024storydiffusion} + I2V \cite{wan2025wan} & 64.27 & \textbf{66.27} & 0.5967 & \textbf{191.3154} & -- & -- \\
CineTrans \cite{wu2025cinetrans}        & 47.27 & 49.95 & 0.6545 & 184.1266 & -- & --  \\
EchoShot \cite{wang2025echoshot}        & 57.36 & 61.11 & 0.6495 & 188.2731 & -- & -- \\
HoloCine \cite{meng2025holocine}        & 54.02 & 55.43 & 0.6667 & 128.1417 & -- & -- \\
\midrule
Phantom \cite{liu2025phantom}           & 75.16 & 55.20 & 0.6136 & 117.6180 & 0.8120 & 18.50\%  \\
VACE \cite{jiang2025vace} (1.3B)        & 65.30 & 48.15 & 0.6043 & 147.2950 & 0.7693 & 22.77\%  \\
VACE \cite{jiang2025vace} (14B)         & 72.10 & 52.40 & 0.5985 & 148.6388 & 0.7597 & 21.68\%  \\
\midrule
\rowcolor{gray!15} 
\textbf{MuSS (Ours)} & \textbf{78.50} & 62.27 & \textbf{0.6990} & 187.8128 & \textbf{0.8827} & \textbf{7.35\%} \\
\bottomrule
\end{tabular}
\end{table*}

\subsection{Experimental Setup}
Our baselines represent a diverse spectrum of current paradigms:
(1) \textit{Storyboard:} StoryDiffusion \cite{zhou2024storydiffusion} combined with Wan2.2-I2V-14B \cite{wan2025wan}, generating auto-regressive keyframes before temporal interpolation. 
(2) \textit{Native Multi-Shot:} CineTrans \cite{wu2025cinetrans}, HoloCine \cite{meng2025holocine}, and EchoShot \cite{wang2025echoshot}, which model long temporal contexts without explicit 2D references. 
(3) \textit{Subject-Driven S2V:} Phantom \cite{liu2025phantom} and VACE \cite{jiang2025vace} (evaluated on 1.3B and 14B), state-of-the-art models conditioned directly on external images. 
(4) \textit{Trivial Copy-Paste Baseline:} A sequence where the 2D reference is pasted onto backgrounds, serving as a physical lower bound to validate our anti-cheating metrics.

\noindent \textbf{Implementation Details.}
Our MuSS-augmented baseline builds on EchoShot.
We perform full-parameter fine-tuning exclusively on MuSS. To condition S2V generation on cross-shot references, we concatenate the reference subject's latent tokens with target multi-shot video latents along the sequence dimension. These concatenated tokens are jointly fed into the Diffusion Transformer self-attention blocks, enabling fine-grained, cross-frame spatiotemporal feature injection.
Training videos are standardized to $832 \times 480$ at 16 fps. We use a multi-shot sliding window of 161 frames. For fair comparison, all methods use prompt extensions following their official guidelines.

\subsection{Track 1: Narrative Effectiveness}

\noindent \textbf{The Absence of Visual-Logic Controllability:} As shown in Table \ref{tab:track1}, concatenation-based methods like StoryDiffusion struggle fundamentally with continuous storytelling, exhibiting a noticeably high Transition Deviation. Although native multi-shot models improve text alignment, they suffer a severe performance drop across the four-dimensional LMM visual logic tests. This performance gap indicates that without rigorous data constraints, models are highly prone to severe structural hallucinations in background environments and frequently break spatial topological rules when the camera perspective switches.

\noindent \textbf{The Superiority of MuSS:} Trained strictly on our dataset, our baseline achieves the most robust overall balance. While HoloCine demonstrates a marginally lower Transition Deviation, and concatenation methods like StoryDiffusion yield a lower Consistency Gap (which is often an artifact of generating unnaturally smooth interpolations between static keyframes rather than authentic temporal dynamics), our MuSS-augmented model consistently dominates all visual logic dimensions. The highly competitive Transition Deviation proves that our model has internalized authentic cinematic editing priors, while the state-of-the-art scores in Scene, Casting, Action, and Spatial Logic confirm that it maintains rigorous structural coherence despite dramatic viewpoint shifts.

\begin{figure*}[t]
\centering
\includegraphics[width=\linewidth]{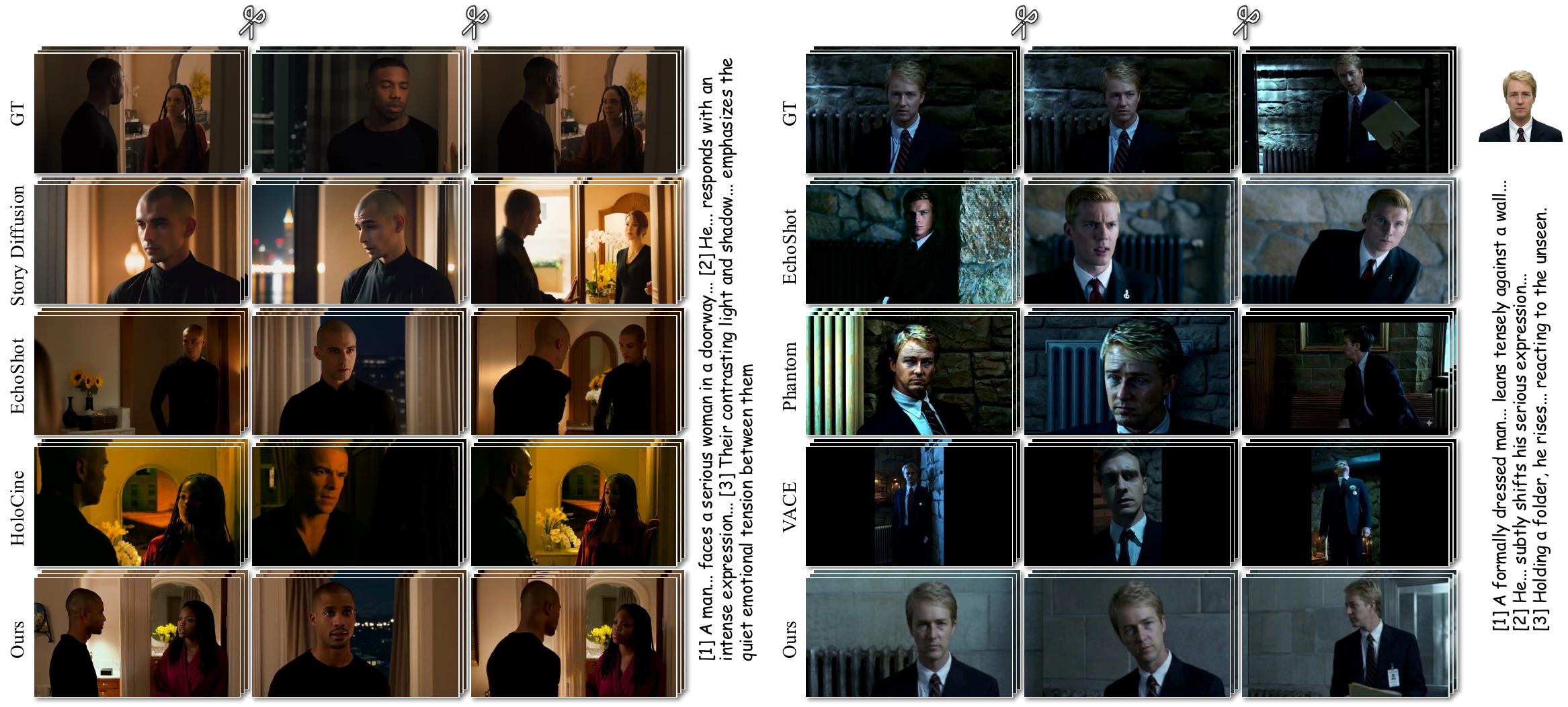}
\caption{Qualitative results on the Cinematic Narrative Benchmark. (Left) Track 1: Evaluating multi-shot consistency across complex cinematic transitions. (Right) Track 2: Assessing 3D identity preservation in Subject-to-Video generation. Our MuSS-augmented baseline effectively resolves severe spatial hallucinations and rigid ``copy-paste'' artifacts of existing methods.}
\Description{Qualitative multi-shot comparisons across several generation methods. The left panels compare scene, casting, action, and spatial continuity through cinematic transitions. The right panels compare subject identity and pose variation, highlighting rigid reference copying in baselines and more varied views from the MuSS-augmented model.}
\label{fig:qualitative}
\end{figure*}

\subsection{Track 2: Subject Consistency}

\noindent \textbf{Measurement Applicability:} Methods like HoloCine, EchoShot, CineTrans, and StoryDiffusion lack external reference inputs. We instead extract the primary subject from their first generated shot as a pseudo-reference for identity evaluation. Consequently, metrics designed to penalize external visual overfitting, namely Anti-Copy-Paste Variance (ACP-Var) and Copy-Paste Rate (CP-Rate), are structurally inapplicable and thus omitted (denoted as ``--'').

\noindent \textbf{Exposing Shortcut Learning:} Table \ref{tab:track2} reveals prevalent ``shortcut learning'' among customizable S2V baselines. Models like Phantom and VACE achieve high Ref-Sub.Con but drastically fail on Inter-Sub.Con, exposing a failure to comprehend intrinsic 3D identity. Coupled with low ACP-Var and high CP-Rate scores, they deteriorate into rigid 2D ``image translation,'' acting as sticker generators that collapse upon perspective shifts.

\noindent \textbf{Effectiveness of Cross-Shot Tracking:} Our baseline successfully severs the pixel-copying shortcut. While native methods like StoryDiffusion maintain internal consistency due to the absence of external reference constraints, our model uniquely solves the S2V challenge. Compared to state-of-the-art customizable baselines, it maintains superior Ref-Sub.Con while significantly bridging the internal consistency gap. Furthermore, yielding the highest ACP-Var and Subject Recall alongside the lowest CP-Rate, empirical results validate that our model masters true 3D structural comprehension and robust subject grounding across complex cinematic camera movements.

\subsection{Qualitative Analysis}

Figure \ref{fig:qualitative} visualizes our benchmark's two evaluation tracks, highlighting the structural limitations of existing models.

\noindent \textbf{Mastering Cinematic Transitions:} During complex narrative transitions, such as an over-the-shoulder shot to a close-up, current baselines fail to maintain the established visual world. StoryDiff and HoloCine exhibit shifts in background architecture and environmental lighting, while EchoShot loses characters. In contrast, our model strictly preserves spatial topology and cast continuity, ensuring authentic cinematic flow.

\noindent \textbf{Breaking the ``Copy-Paste'' Shortcut:} When tasked with generating dynamic actions from a single 2D reference, existing S2V models reveal a severe lack of 3D comprehension. Phantom rigidly projects the reference's frontal pose and bright studio lighting into the target scene, resulting in a jarring, 2D sticker-like clash with the ambient environment. VACE completely collapses the spatial structure and aspect ratio. Empowered by the cross-shot matching mechanism, our model successfully decouples the subject's intrinsic identity from the reference conditions, synthesizing highly natural novel views (e.g., shifting profile, interacting with a folder) perfectly integrated into the target narrative physics.

\begin{table}[t]
\centering
\small
\setlength{\tabcolsep}{4pt}
\caption{Correlation Analysis with Human Judgments. Our proposed metrics demonstrate state-of-the-art alignment with professional human perception.}
\label{tab:correlation}
\begin{tabular}{l l c c}
\toprule
\textbf{Dimension} & \textbf{Core Metric} & \textbf{Spearman's $\rho$} $\uparrow$ & \textbf{Kendall's $\tau$} $\uparrow$ \\
\midrule
\multirow{2}{*}{Narrative Effect.} 
& Trans.Dev  & 0.653 & 0.541 \\
& Scene.Logic  & 0.742 & 0.628 \\
\midrule
\multirow{2}{*}{Subject Consist.} 
& Inter-Sub.Con & 0.618 & 0.505 \\
& \textbf{ACP-Var (Ours)} & \textbf{0.794} & \textbf{0.671} \\
\bottomrule
\end{tabular}
\end{table}

\balance
\subsection{Alignment with Human Perception}

To validate our benchmark's scientific rigor, we conducted a comprehensive blind user study where 15 professional filmmakers independently rated 200 randomly sampled generated sequences on a 1--5 Likert scale. As shown in Table \ref{tab:correlation}, our ACP-Var metric achieves exceptionally high correlation with subjective ratings on motion naturalness and perspective richness, effectively penalizing rigid 2D sticker effects overlooked by traditional metrics. Concurrently, our visual logic metrics (e.g., Scene.Logic) strongly align with expert judgments on visual continuity.

\section{Conclusion}
\label{sec:conclusion}
In this work, we tackle two major bottlenecks hindering video generation: the absence of coherent multi-shot cinematic narratives and the pervasive "copy-paste" shortcut in Subject-to-Video (S2V) synthesis. To overcome these challenges, we present MuSS, a large-scale dataset that leverages a progressive captioning strategy alongside a rigorous cross-shot matching mechanism to guarantee authentic identity preservation. Furthermore, we introduce the Cinematic Narrative Benchmark, equipped with novel LMM-driven Visual Logic and Anti-Copy-Paste Variance (ACP-Var) metrics to systematically evaluate multi-shot generative capabilities. Our extensive experiments demonstrate a stark contrast: while current S2V models structurally deteriorate during dynamic viewpoint shifts, our MuSS-augmented baseline delivers state-of-the-art cross-shot consistency and storytelling effectiveness. Looking ahead, we plan to extend this framework to model complex multi-character interactions and longer narrative arcs with richer temporal dependencies. We believe MuSS will serve as a vital infrastructure, propelling the multimedia community beyond the constraints of isolated single-shot generation and laying the groundwork for robust, industrial-grade cinematic storytelling.

% ...

\clearpage
\bibliographystyle{ACM-Reference-Format}
\bibliography{arxiv}

\clearpage
\appendix
\setcounter{section}{0}
\setcounter{table}{0}
\setcounter{figure}{0}
\renewcommand{\thetable}{S\arabic{table}}
\renewcommand{\thefigure}{S\arabic{figure}}
\section*{Supplementary Material}
\section{Data Source and Access}
\label{sec:data_source}
To construct the MuSS dataset, the raw cinematic video clips were collected from publicly accessible YouTube videos. We do not directly distribute the raw video files. Instead, we provide a comprehensive list of YouTube video IDs alongside the corresponding timestamp annotations for each curated cinematic shot. Copyright and licensing of the source videos remain with their respective providers and rights holders.

To facilitate access and reproducibility for the research community, we include automated download and trimming scripts (\texttt{download\_muss.sh} / \texttt{download\_muss.py}) in the official repository. These scripts leverage open-source tools to fetch source videos using the released identifiers and automatically trim them according to the annotated multi-shot boundaries. Researchers can thereby reproduce the dataset locally; users remain responsible for obtaining and using source content in accordance with applicable platform terms, licenses, intellectual-property rights, and local law.

\section{Extended Dataset Construction Details}
\label{sec:dataset_details}

Building upon the MuSS curation pipeline introduced in Section 3 of the main manuscript, this section provides the specific operational parameters required for reproducibility.

\subsection{Multi-Dimensional Cascaded Filtering Thresholds}
To guarantee that the distilled clips provide a robust foundation for controllable generation, MuSS employs a rigorous cascaded filtering pipeline. It is applied to single-shot candidates before sliding-window sequence construction, with the motion range retaining low-motion dialogue and establishing shots while excluding frozen or excessively chaotic content. The specific empirical thresholds utilized to eliminate severe intra-shot semantic drift, poor visual aesthetics, and inadequate temporal dynamics are summarized in Table~\ref{tab:filtering_thresholds}.

\begin{table}[h]
\centering
\caption{Specific empirical thresholds utilized during the MuSS multi-dimensional filtering curation to ensure robust video quality.}
\label{tab:filtering_thresholds}
\resizebox{0.95\columnwidth}{!}{
\begin{tabular}{llc}
\toprule
\textbf{Filtering Dimension} & \textbf{Core Metric / Model} & \textbf{Threshold} \\
\midrule
Semantic Consistency & CLIP \cite{radford2021learning} / DINO \cite{caron2021emerging} & $\ge 0.80$ \\
Visual Aesthetic Quality & Aesthetic Predictor V2.5 / SigLIP \cite{zhai2023sigmoid} & $\ge 4.00$ \\
Text-Visual Alignment & VideoCLIPXL \cite{xu2021videoclip} & $\ge 0.20$ \\
Textual Describability & Internal Baseline & $\ge 0.02$ \\
Dynamic Motion & Optical Flow (RAFT \cite{teed2020raft}) & $[0.1, 20]$ \\
\bottomrule
\end{tabular}
}
\end{table}

\noindent\textbf{Cross-Shot Reference Construction.}
Subject-centric captions and entity tags first provide candidates for cross-shot association. GroundingDINO and SAM 2.1 masks, together with visual-feature matching, are then used to retain reliable identities. Reference candidates are searched within the same continuous storyline but outside the shots that constitute the target multi-shot clip; pairs are retained at a multimodal text/image similarity threshold of $0.6$ and further checked by the verifier, while ambiguous associations are discarded.

\subsection{Prompt Templates for Progressive Captioning}
A fundamental challenge addressed in the main text is resolving the spatiotemporal text alignment conflict. To ensure full transparency and reproducibility of our Progressive Two-Stage Coherent Captioning strategy, we detail the exact System and User prompt templates deployed to the Large Multimodal Models (LMMs) in Table~\ref{tab:annotation_prompts}.

\begin{table*}[!t]
\centering
\caption{Standardized prompt templates utilized during the MuSS progressive annotation pipeline. This wording is designed to decouple local visual grounding from global narrative coherence.}
\label{tab:annotation_prompts}
\begin{tabular}{@{} p{0.17\linewidth} p{0.10\linewidth} p{0.67\linewidth} @{}}
\toprule
\textbf{Stage} & \textbf{Role} & \textbf{Prompt Template} \\
\midrule
\multirow{2}{*}{\shortstack[l]{Single-Shot\\Captioning}}
& System
& You are a film-director assistant. Describe a single physical shot from a movie in a detailed, visually grounded, and cinematically useful manner. Focus only on directly visible content. Describe the main subject, actions, appearance, attributes, spatial layout, background elements, lighting, weather, time-of-day cues, and camera-relevant information when such cues are visually evident. Do not invent backstory, motivation, or emotional interpretation not supported by the visual evidence. Avoid generic openings such as ``this video shows''. Prefer concrete noun phrases and clear action verbs. \\
\cmidrule{2-3}
& User
& Please describe this shot in detail. The description should emphasize the visible subject, actions, attributes, scene context, and any camera or visual-style cues that are clearly present. Do not repeat content, and do not infer facts that are not visually grounded. \\
\midrule
\multirow{2}{*}{\shortstack[l]{Caption\\Rewrite}}
& System
& You are a prompt rewriting assistant for video generation. Rewrite the input description so that it is maximally useful for regenerating the same visual content. Preserve only visually supported content. Remove subjective interpretation, unsupported inference, redundant lead-in phrases, and non-existent details. Keep the rewritten description focused on the main subject, actions, attributes, background, location, weather, time, and camera or style cues. Never add new facts. Return a JSON object: \texttt{\{"rewritten description": "..."\}}. \\
\cmidrule{2-3}
& User
& Rewrite the following shot description into a concise, visually grounded, generation-oriented description. Avoid subjective wording and do not add information that is not explicitly supported by the visual content. \\
\midrule
\multirow{2}{*}{\shortstack[l]{Multi-Shot\\Coherent Aggregation}}
& System
& You are a film-director assistant responsible for refining a sequence of consecutive movie shots into a coherent multi-shot description. Each shot already has an initial local caption. Preserve the local visual facts of every shot while improving sequence-level coherence. Introduce each character, object, or location only when it first appears. In later shots, maintain stable references using clear pronouns or concise descriptors. Resolve contradictions across shots. Output one caption per shot in the format ``Shot 1: ...'', ``Shot 2: ...''. Do not merge shots, omit shots, or reorder them. \\
\cmidrule{2-3}
& User
& The following shots belong to the same continuous scene or sequence. Each shot is accompanied by an initial local description. Please refine all captions jointly so that the full sequence is narratively coherent, while preserving shot-level visual accuracy. Ensure consistent entity references across shots and return exactly one caption per shot. \\
\bottomrule
\end{tabular}
\end{table*}

\section{Extended Benchmark Implementation Details}
\label{sec:extended_benchmark}

To ensure complete transparency and reproducibility of our Cinematic Narrative Benchmark, we detail the specific Large Multimodal Model (LMM) prompts utilized for the Multi-Dimensional Visual Logic (MDVL) evaluation. The benchmark prompts are selected from the MuSS test set, balanced across the two tracks, shot count, indoor/outdoor scene type, dialogue/action rhythm, close-up/wide composition, and single-/multi-subject configurations, and manually inspected after prompt writing and stratified selection.

As introduced in the main manuscript, the reported MDVL scores use Gemini-2.5-Flash \cite{geminiteam2025gemini25} to evaluate four crucial dimensions of cinematic continuity. To achieve reliable and interpretable scoring, we construct a $2 \times N$ visual grid by uniformly sampling 2 keyframes from each of the $N$ generated sub-shots. This visual grid is fed into Gemini-2.5-Flash alongside the global narrative description and the local shot prompts. The standardized prompt template used to enforce strict visually grounded reasoning is presented in Table~\ref{tab:mdvl_prompt}.

\begin{table*}[!t]
\centering
\caption{The standardized System Prompt utilized for the Multi-Dimensional Visual Logic (MDVL) evaluation. The LMM evaluates the generated $2 \times N$ visual grid strictly based on the provided cinematic definitions.}
\label{tab:mdvl_prompt}
\begin{tabular}{@{} p{0.95\linewidth} @{}}
\toprule
\textbf{MDVL Evaluation Prompt Template} \\
\midrule
You are a professional film visual narrative reviewer. \\
Below is a keyframe grid of a generated multi-shot video. Each column corresponds to a sub-shot, containing 2 uniformly sampled keyframes (in chronological order from top to bottom). The columns from left to right are Shot 1, Shot 2, Shot 3, etc. \\
\\
\textbf{Global Narrative Description:} \{global\_narrative\} \\
\textbf{Local Prompts for Each Shot:} \{shot\_prompts\} \\
\\
Please strictly base your evaluation on the \textbf{visual content} (do not guess merely from the text) and score the following 4 dimensions on a scale of 1 to 5. \\
\\
\textbf{Dimension 1: Scene.Logic (Scene Consistency)} \\
Evaluate the consistency of the background environment after cuts. 
- When cutting to a different scene and back, do the background details (furniture, lighting, windows, plants, etc.) remain identical? 
- Score 5 = Background details are perfectly consistent, seamless cuts. | Score 1 = Background changes every shot, completely chaotic. \\
\\
\textbf{Dimension 2: Casting.Logic (Identity Consistency)} \\
Evaluate the preservation of character identities across shots. 
- Do the physical traits (hairstyle, clothing color, body type) of the same character remain consistent? 
- Note: Do not penalize if a character is omitted in certain shots due to framing/perspective. 
- Score 5 = All identities perfectly maintained. | Score 1 = Severe arbitrary mutations (e.g., face swapping, outfit changing). \\
\\
\textbf{Dimension 3: Act.Logic (Action Continuity)} \\
Evaluate the temporal continuation of dynamic behaviors. 
- Do dynamic actions (talking, walking, interacting) form a logical temporal continuation after the cut? 
- Score 5 = Fluent action transitions, temporally coherent. | Score 1 = Complete jump-cut, zero action linkage logic. \\
\\
\textbf{Dimension 4: Spat.Logic (Spatial Topology)} \\
Evaluate the spatial relationships across shots. 
- Are relative positions maintained reasonably? (e.g., if A is on the left and B is on the right, they should not teleport). Does it follow the 180-degree rule? 
- Score 5 = Spatial relationships are completely logical. | Score 1 = Character positions are entirely messed up. \\
\\
Please output a single JSON object in the following format: \\
\texttt{\{ "scene\_logic": <1-5>, "casting\_logic": <1-5>, "act\_logic": <1-5>, "spat\_logic": <1-5>, "reasoning": "<concise analysis>" \}} \\
\bottomrule
\end{tabular}
\end{table*}

\section{Implementation \& Training Details}
\label{sec:implementation}
As detailed in Section 5.1 of the main text, our MuSS-augmented baseline is constructed upon the EchoShot framework \cite{wang2025echoshot} using latent sequence concatenation to inject identity priors. This section provides the precise training hyperparameters utilized to achieve convergence:
\begin{itemize}
    \item \textbf{Optimizer}: AdamW ($\beta_1 = 0.9$, $\beta_2 = 0.999$, weight decay $= 10^{-4}$).
    \item \textbf{Learning Rate}: $1 \times 10^{-5}$ with a linear warmup of 2,000 steps.
    \item \textbf{Total Training Steps}: 50,000.
    \item \textbf{Resolution \& Framerate}: $832 \times 480$ spatial resolution at 16 fps.
    \item \textbf{Temporal Context}: 161 frames processed via a multi-shot sliding-window approach.
\end{itemize}
Training was executed on 32 NVIDIA H20 GPUs, requiring approximately 3.5 days to reach convergence. Reference preparation is handled by the automated curation pipeline, while benchmark evaluation additionally requires the domain-specific preprocessing and LMM inference described above for each evaluated sequence.

\section{User Study Details}
\label{sec:user_study}
To empirically validate that our Visual-Logic Driven benchmark aligns with professional human perception (refer to Section 5.5 in the main text), we conducted a rigorous blind user study. We recruited 15 professional filmmakers (directors, editors, and cinematographers), each possessing a minimum of three years of industry experience. Participants independently evaluated 200 randomly sampled generated sequences completely blind to the generating model. The evaluation criteria encompassed temporal naturalness, narrative coherence, visual continuity across cuts, motion naturalness, and overall subject consistency.

\paragraph{Evaluation Rubric.} To ensure the subjective scores directly correlate with industrial production standards, the expert evaluators were instructed to rate the generated sequences using the following strict 1--5 Likert scale rubric:
\begin{itemize}
    \item \textbf{5 - Cinematic Grade:} The sequence exhibits flawless temporal continuity, robust identity preservation, and logical spatial transitions indistinguishable from professional editing.
    \item \textbf{4 - High Quality:} Minor artifacts may exist, but the core narrative logic, subject identity, and scene structure are well-maintained across cuts.
    \item \textbf{3 - Acceptable:} The sequence follows the general prompt, but exhibits noticeable inconsistencies in background details or minor identity drift (e.g., clothing color changes).
    \item \textbf{2 - Poor Logic:} Severe jump-cuts, broken spatial topology, or obvious 2D "copy-paste" artifacts that disrupt the viewing experience.
    \item \textbf{1 - Complete Failure:} The sequence lacks any multi-shot logic, subjects mutate arbitrarily, or the video degrades into static slideshows.
\end{itemize}

\section{Extended Dataset Visualizations and Limitations}
\label{sec:dataset_vis}

\subsection{More Visualizations of the Dataset}
To provide a deeper understanding of the MuSS dataset's scale, diversity, and visual fidelity, we present extended visualizations of the curated clips. The dataset spans a wide array of cinematic genres, effectively capturing complex lighting environments, varied spatial layouts, and dynamic subject motions. 

\begin{figure*}[t]
    \centering
    \includegraphics[width=0.95\linewidth,height=0.78\textheight,keepaspectratio]{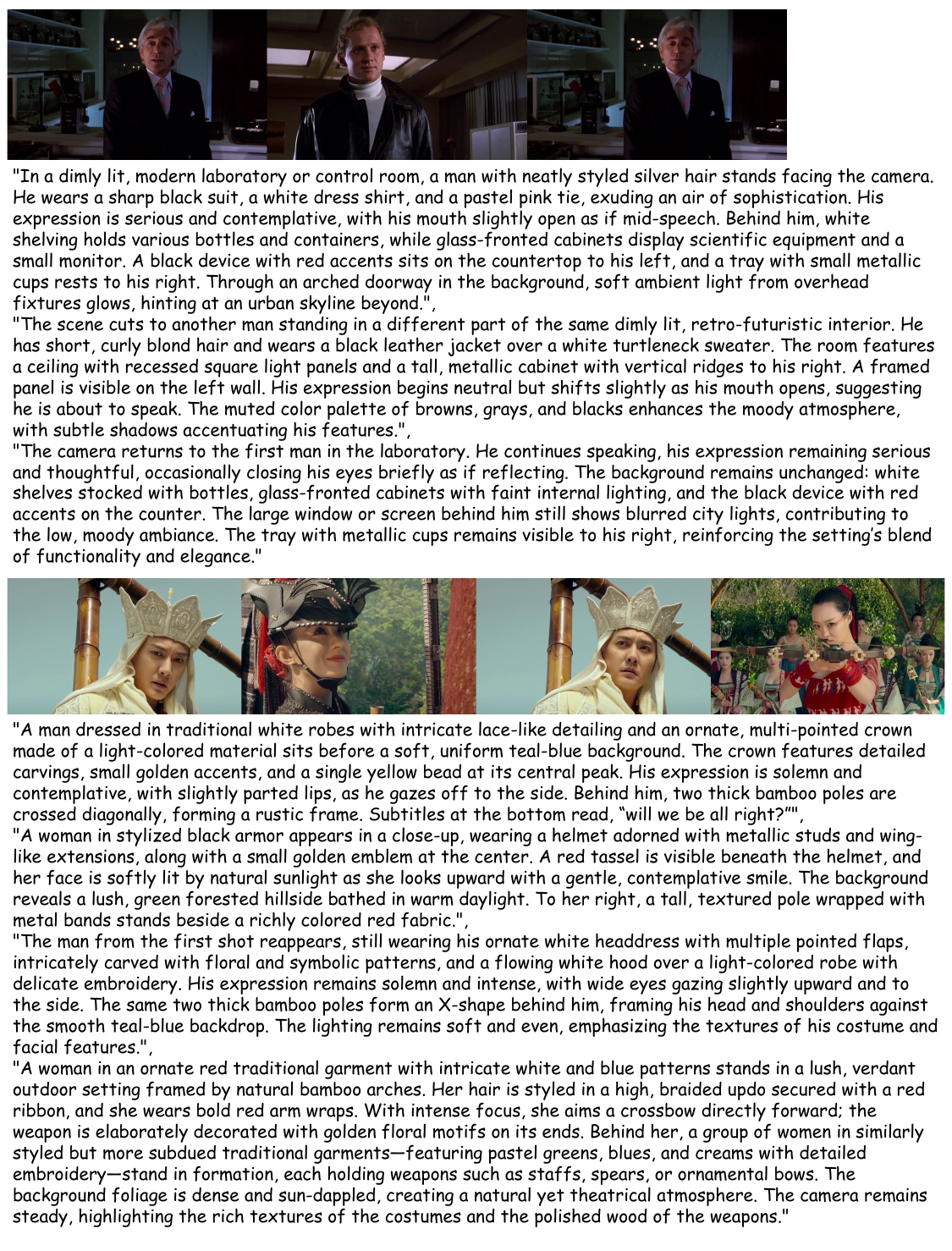}
    \caption{Data examples for Track 1 (Complex Cinematic Narratives). This visualization showcases the curated keyframes paired with their corresponding progressive multi-shot captions, demonstrating the precise spatiotemporal text alignment achieved by our annotation pipeline.}
    \Description{Examples of complex cinematic narratives showing multi-shot keyframes together with their progressively constructed shot-level captions.}
    \label{fig:suppl_vis_track1_data}
\end{figure*}

\begin{figure*}[t]
    \centering
    \includegraphics[width=0.9\linewidth,height=0.78\textheight,keepaspectratio]{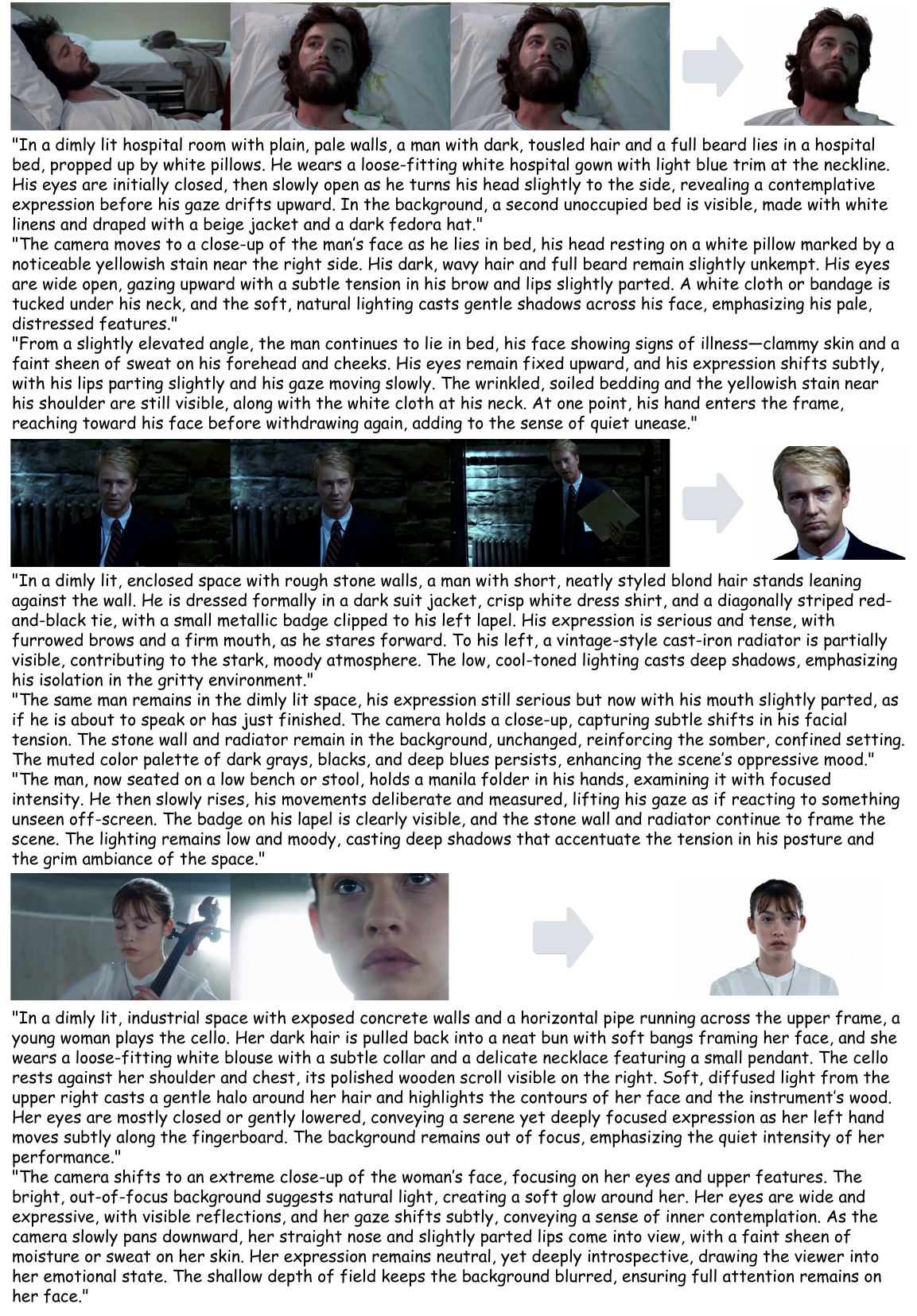}
    \caption{Data examples for Track 2 (Subject-Centric Narratives). This figure details the data structure by presenting the isolated reference subject alongside dynamic target shots and their structured captions, which explicitly force models to learn novel-view synthesis rather than pixel copying.}
    \Description{Subject-centric data examples pairing an isolated reference subject with target shots from different views and their structured captions.}
    \label{fig:suppl_vis_track2_data}
\end{figure*}

\begin{figure*}[t]
    \centering
    \includegraphics[width=0.95\linewidth,height=0.78\textheight,keepaspectratio]{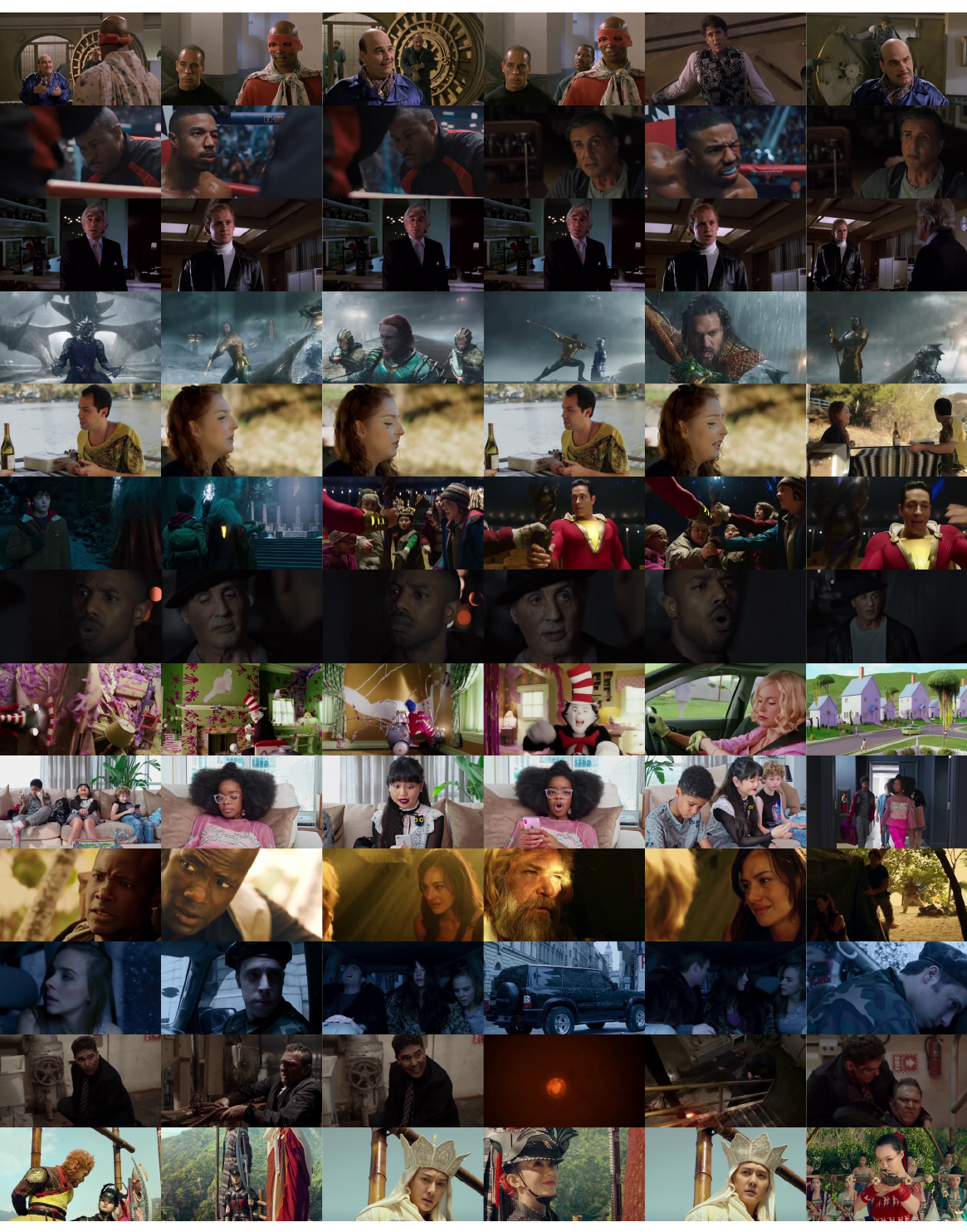}
    \caption{Raw cinematic transitions for Track 1 (Complex Cinematic Narratives). These unannotated screenshot sequences highlight the intricate montage transitions across diverse cinematic genres, showcasing the dataset's capacity to model structural logic in professional editing.}
    \Description{Raw screenshot sequences illustrating diverse cinematic edits, including changes in framing, viewpoint, setting, and subject composition.}
    \label{fig:suppl_vis_track1_raw}
\end{figure*}

\begin{figure*}[t]
    \centering
    \includegraphics[width=0.95\linewidth,height=0.78\textheight,keepaspectratio]{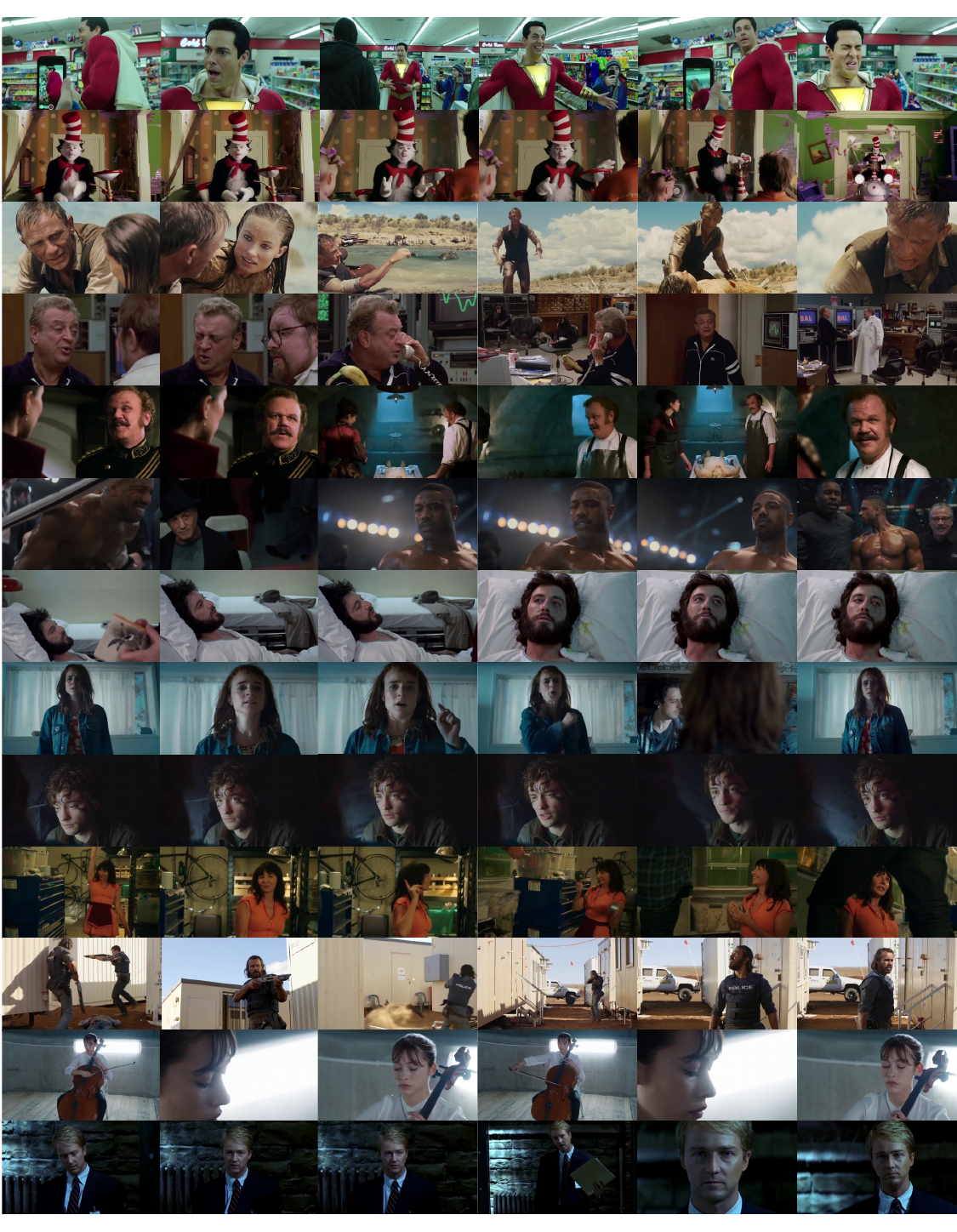}
    \caption{Raw cinematic sequences for Track 2 (Subject-Centric Narratives). These extensive multi-shot screenshots demonstrate robust cross-shot subject consistency in authentic film data, capturing identical subjects across varying camera angles, lighting conditions, and backgrounds.}
    \Description{Raw multi-shot sequences showing recurring subjects across changes in camera angle, pose, lighting, and background.}
    \label{fig:suppl_vis_track2_raw}
\end{figure*}

Specifically, we present four detailed visualizations to highlight our two core narrative tracks:
\begin{itemize}
    \item \textbf{Complex Cinematic Narratives (Figures~\ref{fig:suppl_vis_track1_data} and \ref{fig:suppl_vis_track1_raw}):} We showcase intricate montage transitions, such as cutting from an establishing wide shot to an over-the-shoulder dialogue, and shifting between multiple characters within the same continuous scene. Figure~\ref{fig:suppl_vis_track1_data} displays curated data examples including keyframes and their corresponding progressive captions, while Figure~\ref{fig:suppl_vis_track1_raw} provides a broader view of raw cinematic screenshot sequences.
    \item \textbf{Subject-Centric Narratives (Figures~\ref{fig:suppl_vis_track2_data} and \ref{fig:suppl_vis_track2_raw}):} We display a core reference subject alongside a diverse set of multi-view target shots, illustrating how the identical identity is captured across drastically different camera angles, postures, and background contexts. Figure~\ref{fig:suppl_vis_track2_data} details the annotated data format with extracted subjects, target keyframes, and captions, whereas Figure~\ref{fig:suppl_vis_track2_raw} presents extensive raw multi-shot cinematic screenshots of consistent subjects.
\end{itemize}

\subsection{Limitations and Future Work}
While MuSS substantially mitigates the copy-paste shortcut and pioneers logical evaluation, several challenges remain. Among 86 difficult cases identified during curation and audit, the main sources of error were heavy occlusion or crowds ($31\%$), extreme close-ups or viewpoint changes ($24\%$), low-light conditions or motion blur ($22\%$), ambiguous multi-character identity switches ($19\%$), and formatting or tool errors ($4\%$). These cases can perturb the zero-shot subject extraction pipeline, while micro-level identity features (e.g., intricate reflective jewelry or fine textures) may also exhibit temporal instability under drastic multi-shot perspective shifts. Finally, our benchmark's reliance on state-of-the-art LMMs inherits their inherent limitations regarding extremely fine-grained causal reasoning. Future iterations will focus on explicitly modeling 3D-aware priors and expanding the dataset to encompass complex multi-subject interactive narratives.

\section{Ethical Considerations}
\label{sec:ethics}
MuSS is developed exclusively to advance academic research in multi-shot cinematic storytelling and controllable Subject-to-Video generation. By explicitly designing data curation and evaluation metrics that penalize superficial visual copying, we aim to steer the community toward structurally grounded, physics-aware synthesis. However, we acknowledge the dual-use nature of highly consistent subject-driven generation, which poses risks of deepfake generation or deceptive synthesis. We strongly advocate that future public deployment of such robust S2V frameworks be strictly coupled with invisible watermarking technologies (e.g., SynthID), comprehensive generation logging, and robust provenance tracking to ensure responsible innovation in multimedia content creation.

\end{document}